\documentclass{article}
\usepackage{spconf,amsmath,graphicx,float}

\usepackage{booktabs}
\usepackage{xcolor}
\usepackage{pifont}
\usepackage{wrapfig}
\usepackage{url}
\usepackage{multirow}
\usepackage{algorithm}
\usepackage{algorithmic}
\usepackage{caption}
\usepackage{amssymb}
\usepackage{cite}

\usepackage{listings}
\usepackage{xcolor}

\lstdefinelanguage{yaml}{
  keywords={concept_token, subject, style, settings},
  keywordstyle=\color{blue}\bfseries,
  basicstyle=\ttfamily\small,
  commentstyle=\color{gray},
  stringstyle=\color{teal},
  sensitive=false
}

\title{ReDiStory: Region-Disentangled Diffusion for Consistent Visual Story Generation}
\name{Ayushman Sarkar$^{1}$, Zhenyu Yu$^{2,*}$, Chu Chen$^{3}$, Wei Tang$^{3}$, Kangning Cui$^{3}$, Mohd Yamani Idna Idris$^{2}$}
\address{
${^1}$ Birbhum Institute of Engineering and Technology,
${^2}$ Universiti Malaya, \\
${^3}$ City University of Hong Kong, 
${^4}$ Wake Forest University%\\
}

\begin{document}
%\ninept
%
\maketitle

% fig_motivation
% \twocolumn[{
% \renewcommand\twocolumn[1][]{#1}
% \maketitle
% \begin{center}
%     % \vspace{-10pt} %15
%     \captionsetup{type=figure}
%     \includegraphics[width=0.87\linewidth]{figs/fig_motivation1.png}
%     \captionsetup{type=figure, width=1.0\linewidth} % 调整 caption 的宽度和图片一致
%     % \vspace{-15pt} %15
%     % \caption{Motivation.}
%     % \caption{Motivation. Prompt concatenation in training-free multi-frame storytelling often entangles identity and frame-specific semantics, causing inter-frame semantic interference and inconsistent subject appearance. ReDiStory mitigates this issue by reorganizing prompt embeddings to better preserve shared identity across frames.}
%     % \caption{Motivation. Prompt concatenation in training-free multi-frame storytelling can entangle identity and frame-specific semantics, leading to inconsistent subject appearance. ReDiStory addresses this issue through prompt embedding reorganization.}
%     \caption{Motivation. Concatenating an identity prompt with multiple frame prompts can cause semantic interference that weakens identity cues and leads to inconsistent subject appearance. ReDiStory mitigates this issue by disentangling identity and frame-specific embeddings and decorrelating frame representations at inference time.}
%     \label{fig_motivation}
% \end{center}
% }]

%
% \vspace{-30pt}
\begin{abstract}
Generating coherent visual stories requires maintaining subject identity across multiple images while preserving frame-specific semantics. Recent training-free methods concatenate identity and frame prompts into a unified representation, but this often introduces inter-frame semantic interference that weakens identity preservation in complex stories. We propose \textit{ReDiStory}, a training-free framework that improves multi-frame story generation via inference-time prompt embedding reorganization. \textit{ReDiStory} explicitly decomposes text embeddings into identity-related and frame-specific components, then decorrelates frame embeddings by suppressing shared directions across frames, reducing cross-frame interference without modifying diffusion parameters or requiring additional supervision. Under identical diffusion backbones and inference settings, \textit{ReDiStory} improves identity consistency while maintaining prompt fidelity. Experiments on the ConsiStory+ benchmark show consistent gains over 1Prompt1Story on multiple identity consistency metrics. Code is available at: https://github.com/YuZhenyuLindy/ReDiStory
\end{abstract}
% On the ConsiStory+ benchmark, ReDiStory consistently outperforms 1Prompt1Story on multiple identity consistency metrics with negligible overhead.

\section{Introduction}
\label{sec:intro}

Text-to-image (T2I) diffusion models~\cite{rombach2022high,saharia2022photorealistic,yu2025visualizing} can synthesize high-quality images from textual descriptions and have become a standard foundation for controllable image generation \cite{gaur2025storysync,ma2025lay2story,yu2025cotextor}. Beyond single-image generation, \emph{visual storytelling} aims to generate a sequence of images that repeatedly depicts the same subject across varying scenes and actions \cite{yang2024seed}. This capability is essential for illustrated storybooks, animation pre-visualization, and sequential content creation~\cite{hu2024animate,yang2024seed}.

\begin{figure}[t]
    \centering
    \includegraphics[width=1.0\linewidth]{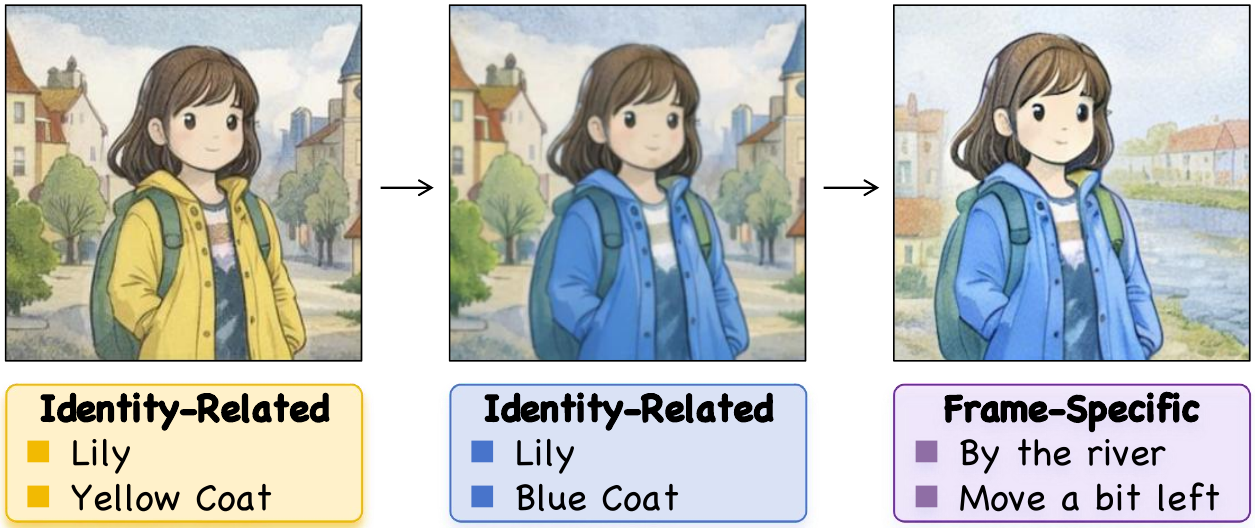}
    \caption{\textbf{Motivation.} Identity-related components (e.g., name and appearance) can be entangled with frame-specific instructions (e.g., location and pose), leading to unintended identity changes across frames. We therefore separate these two components for more consistent multi-frame generation.}
    \label{fig:motivation_components}
\end{figure}

As illustrated in Fig.~\ref{fig:motivation_components}, a key source of identity drift is the entanglement between identity-related attributes and frame-specific instructions in prompt representations. A central challenge is to preserve \emph{subject identity} across frames without sacrificing \emph{frame-specific semantics}. When frames are generated independently, diffusion models often exhibit identity drift and attribute inconsistency. Existing solutions address this issue either by subject-driven fine-tuning~\cite{ruiz2023dreambooth,gal2022textual}, which requires retraining and additional data, or by training-free methods that manipulate attention or prompt structure at inference time~\cite{tewel2024training,hertz2022prompt}. The latter offers a lightweight and model-agnostic alternative that is more suitable for practical multi-frame generation.

Among training-free approaches, 1Prompt1Story~\cite{liu2025one} concatenates the identity prompt and all frame prompts into a single input sequence, leveraging the contextual coherence of text encoders. While effective, this concatenation strategy also introduces \emph{inter-frame semantic interference}: when frame prompts are encoded jointly, their representations can become correlated in the embedding space; as a result, frame-specific semantics may partially override shared identity cues, leading to inconsistent subject appearance across frames.

We propose \textit{ReDiStory}, a training-free prompt embedding reorganization framework that mitigates inter-frame semantic interference. \textit{ReDiStory} operates purely on text embeddings at inference time. It first decomposes prompt embeddings into identity-related and frame-specific components, and then reorganizes frame embeddings to suppress cross-frame correlation without modifying the diffusion model or introducing additional supervision. Under identical diffusion backbones and inference settings, \textit{ReDiStory} improves identity consistency while maintaining prompt fidelity, yielding consistent gains over the baselines.
Our contributions are three-fold:
\begin{itemize}
    \item We identify \emph{inter-frame semantic interference} as a key reason for identity drift in multi-frame generation.
    \item We propose \textit{ReDiStory}, an inference-time prompt embedding reorganization framework that decouples identity and frame-specific semantics without retraining.
    \item We evaluate \textit{ReDiStory} on the ConsiStory+ benchmark under identical settings  and show improved identity consistency without degrading prompt fidelity.
\end{itemize}

\section{Related Work}
\label{sec:related}

Visual storytelling requires maintaining a persistent subject identity while allowing large, frame-specific scene changes. Existing solutions can be roughly grouped by \emph{where} they enforce consistency: (i) by adapting the model or injecting extra identity cues (strong identity but heavier assumptions), (ii) by keeping the model fixed and coordinating multi-frame generation at inference time (lightweight but prone to drift), and (iii) by relying on prompt design, where concatenation can unintentionally couple frame semantics and cause identity cues to be overridden. We review these three lines to motivate our focus on a purely prompt-level, training-free mitigation of cross-frame semantic interference.

\subsection{Subject-Consistent Text-to-Image Generation.}
Maintaining subject consistency across multiple generated images has been widely studied in text-to-image generation \cite{yu2025cotextor,yu2025visualizing}. Training-based personalization methods, such as DreamBooth~\cite{ruiz2023dreambooth} and Textual Inversion~\cite{gal2022textual}, adapt diffusion models to specific subjects using additional data. These methods typically offer strong identity preservation, but require per-subject optimization and may reduced flexibility when a story involves diverse scenes, poses, or attributes. Recent approaches, including IP-Adapter~\cite{ye2023ip} and PhotoMaker~\cite{li2024photomaker}, use reference images to guide identity customization, but still rely on auxiliary inputs or model adaptation.

\subsection{Training-Free Multi-Frame Storytelling.}
To avoid retraining, training-free approaches operate purely at inference time and are thus well suited for multi-frame storytelling with off-the-shelf diffusion models. ConsiStory~\cite{tewel2024training} propagates subject information through shared attention across batched generations, while StoryDiffusion~\cite{zhou2024storydiffusion} introduces consistent self-attention to stabilize long-range narrative generation. Most closely related to our work, 1Prompt1Story~\cite{liu2025one} concatenates the identity and all frame prompts into a single input sequence, and enhances consistency via embedding reweighting and identity-preserving attention, while keeping the diffusion model unchanged.

\subsection{Prompt Structure and Semantic Interference.}
Beyond model- or attention-level modifications, recent studies highlight that prompt structure can substantially influence generation behavior~\cite{hertz2022prompt}. Concatenation-based prompting can exploit the contextual coherence of language encoders, but may also couple frame representations in the embedding space. In multi-frame settings, such coupling can induce \emph{inter-frame semantic interference}, where frame-specific descriptions partially override shared identity attributes and lead to identity drift across frames~\cite{liu2025one,tewel2024training}. \textit{ReDiStory} targets this prompt-related case by reorganizing frame-level prompt embeddings at inference time, mitigating inter-frame semantic interference without explicit state representations, additional supervision, or changes to the underlying diffusion model.
\begin{algorithm}[t]
% \caption{ReDiStory: Inference-time Prompt Embedding Reorganization}
\caption{\textit{ReDiStory}: Prompt Embedding Reorganization}
\label{alg:redistory}
\begin{algorithmic}[1]
\REQUIRE Identity prompt $P_{id}$, frame prompts $\{P_n\}_{n=1}^{N}$
\REQUIRE Text encoder $\mathcal{E}$, diffusion model $\mathcal{G}_\theta$
\FOR{$n = 1$ to $N$}
    \STATE $E_n \leftarrow \mathcal{E}(P_{id} \oplus P_n)$
    \STATE $E^{id} \leftarrow \mathrm{Slice}_{id}(E_n), \;\; E_n^{f} \leftarrow \mathrm{Slice}_{f}(E_n)$
\ENDFOR
\FOR{$n = 1$ to $N$}
    \STATE $\tilde{E}_n^{f} \leftarrow E_n^{f} - \frac{1}{N-1}\sum_{m \neq n}\mathrm{Proj}_{E_m^{f}}(E_n^{f})$
    \STATE $\tilde{E}_n \leftarrow [E^{id}; \tilde{E}_n^{f}]$
\ENDFOR
\FOR{$n = 1$ to $N$}
    \STATE $I_n \sim \mathrm{Sample}(\mathcal{G}_\theta \mid \tilde{E}_n)$
\ENDFOR
\RETURN $\{I_n\}_{n=1}^{N}$
\end{algorithmic}
\end{algorithm}

\section{Method}
\label{sec:method}

This section introduces \textit{ReDiStory}, motivated by the observation that jointly encoding identity and frame prompts can entangle persistent identity cues with frame-specific instructions, leading to cross-frame semantic interference and identity drift. Our key idea is to explicitly separate identity-related embeddings from frame-specific embeddings, and then remove directions that are shared across frames from the frame-specific part. This inference-time reorganization requires no additional supervision or parameter updates. Algorithm~\ref{alg:redistory} provides an end-to-end summary of the overall pipeline.

\subsection{Problem Formulation}
\label{sec:problem}

We consider the task of multi-frame visual story generation under a training-free setting.
Given an identity prompt $P_{id}$ that describes the subject and a set of $N$ frame prompts
$\{P_n\}_{n=1}^{N}$ specifying frame-level variations, the goal is to generate a sequence of images
$\{I_n\}_{n=1}^{N}$ such that the subject identity remains consistent across frames while each image
faithfully follows its corresponding frame prompt.

Let $\mathcal{G}_\theta$ denote a pretrained text-to-image diffusion model with parameters $\theta$.
Under the training-free setting, frames are generated independently at inference time:
\begin{equation}
I_n \sim \mathcal{G}_\theta(P_{id}, P_n), \quad n = 1, \dots, N.
\end{equation}
In training-free multi-frame generation, a key challenge lies in preserving identity-related semantics
across frames without modifying $\theta$ or introducing additional supervision.
Prior work has shown that naive prompt concatenation often leads to \emph{inter-frame semantic interference},
where frame-specific content suppresses or distorts shared identity attributes.
\textit{ReDiStory} addresses this issue by reorganizing prompt embeddings before they are consumed by the diffusion model.

% --------------------------------------------------

\begin{figure}
    \centering
    \includegraphics[width=1\linewidth]{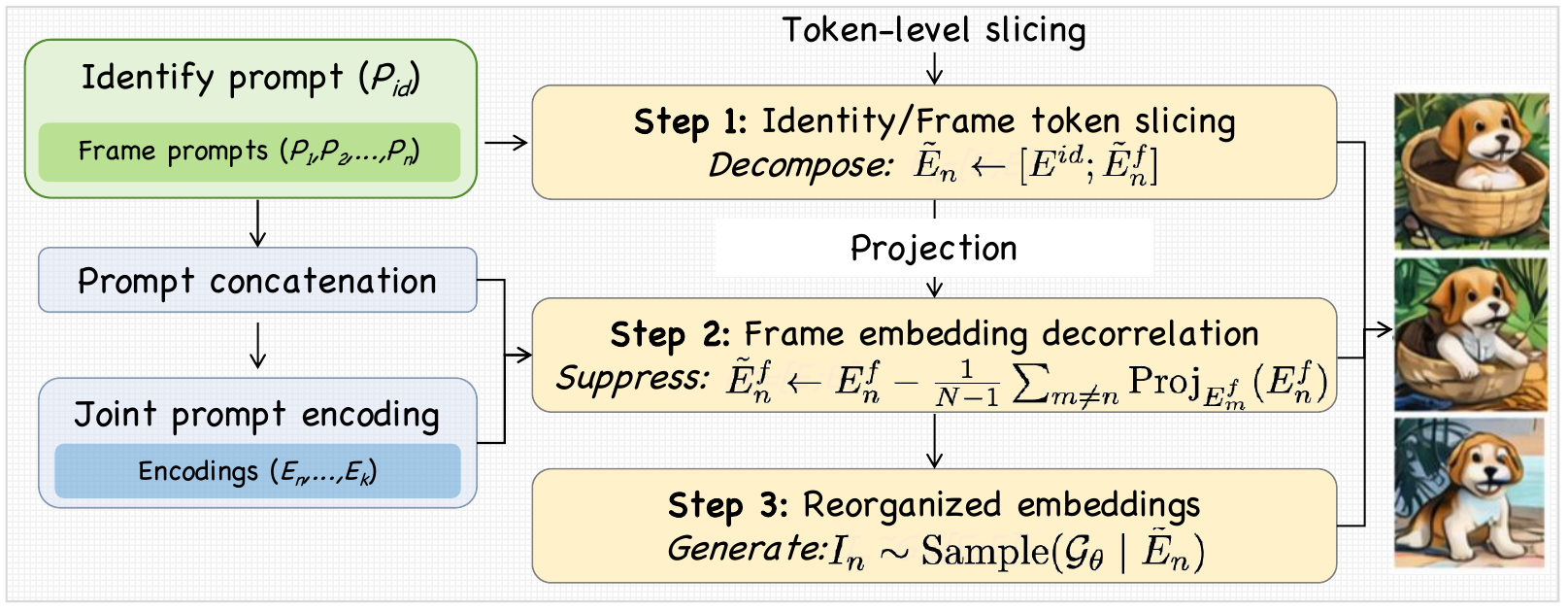}
    % \caption{Overview}
    \caption{ReDiStory pipeline. Identity and frame prompts are jointly encoded, decomposed at the token level, and reorganized by projection-based decorrelation to reduce inter-frame semantic interference before standard diffusion sampling.}
    \label{fig:overview}
\end{figure}

\subsection{Prompt Embedding Representation}
\label{sec:representation}

We first describe how prompt embeddings are represented and decomposed in \textit{ReDiStory}. Following 1Prompt1Story~\cite{liu2025one}, we concatenate an identity prompt $P_{id}$ with a frame prompt $P_n$ and encode them with a text encoder $\mathcal{E}(\cdot)$:
\begin{equation}
E_n = \mathcal{E}(P_{id} \oplus P_n) \in \mathbb{R}^{L_n \times d},
\end{equation}
where $L_n$ denotes the token length and $d$ is the embedding dimension. Because $P_{id}$ and $P_n$ are encoded jointly, identity cues may become entangled with frame-specific instructions in the shared embedding space.

\textit{ReDiStory} makes this structure explicit by partitioning $E_n$ into identity-related and frame-specific parts:
\begin{equation}
E_n = \left[ E^{id}; E_n^{f} \right],
\end{equation}
where $E^{id} \in \mathbb{R}^{L_{id} \times d}$ corresponds to token embeddings of $P_{id}$ (shared across frames) and $E_n^{f} \in \mathbb{R}^{L_n^{f} \times d}$ corresponds to those of $P_n$.
This decomposition enables asymmetric treatment of shared identity and frame-level content, which forms the basis of the proposed reorganization strategy.
We denote this extraction by $E^{id}=\mathrm{Slice}_{id}(E_n)$ and $E_n^{f}=\mathrm{Slice}_{f}(E_n)$, which simply select the embedding rows aligned with identity tokens and frame tokens, respectively. Since the identity prompt is identical across frames, $E^{id}$ can be consistently extracted from any $E_n$.

% --------------------------------------------------

\subsection{Prompt Embedding Reorganization}
\label{sec:reorganization}

To mitigate inter-frame semantic interference, \textit{ReDiStory} reorganizes frame-specific embeddings by removing directions that are shared across frames. Let $\{E_n^{f}\}_{n=1}^{N}$ denote the set of frame-specific embedding matrices. For each frame $n$, we compute a reorganized embedding $\tilde{E}_n^{f}$ as
\begin{equation}
\tilde{E}_n^{f}
=
E_n^{f}
-
\frac{1}{N-1}
\sum_{m \neq n}
\mathrm{Proj}_{E_m^{f}}(E_n^{f}),
\end{equation}
% where $\mathrm{Proj}_{A}(B)$ denotes the projection of $B$ onto the subspace spanned by $A$.
where $\mathrm{Proj}_{A}(B)$ projects $B$ onto the subspace spanned by $A$ in the embedding space (e.g., the column space of $A$), and is applied row-wise to the token embeddings in $B$. Intuitively, this subtraction suppresses cross-frame shared directions, encouraging $\tilde{E}_n^{f}$ to retain frame-specific information rather than drift toward other frames.

% This operation suppresses semantic directions that are shared across frames, encouraging $\tilde{E}_n^{f}$ to focus on frame-specific information.
We then form the reorganized prompt embedding for frame $n$ as
\begin{equation}
\tilde{E}_n = \left[ E^{id}; \tilde{E}_n^{f} \right].
\end{equation}
This procedure is deterministic and performed entirely at inference time, without any additional objectives, gradients, or parameter updates.
% --------------------------------------------------

\subsection{Generation and Computational Complexity}
\label{sec:generation}

Given $\tilde{E}_n$, we generate each frame using the same diffusion sampler as the base model.
Let $z_t^{(n)}$ denote the latent of frame $n$ at diffusion step $t$, where $t=T,\dots,1$ with $T$ the total number of denoising steps. Conditioned on the reorganized text embedding $\tilde{E}_n$, the sampler iterates:
\begin{equation}
z_{t-1}^{(n)} \sim p_\theta \left( z_{t-1}^{(n)} \mid z_t^{(n)}, \tilde{E}_n \right),
\quad t = T, \dots, 1,
\end{equation}
where $p_\theta$ denotes the diffusion model's conditional denoising transition (i.e., the sampling kernel) parameterized by $\theta$. From the perspective of the diffusion model, \textit{ReDiStory} does not alter the architecture or inference procedure, but provides reorganized prompt embeddings that reduce cross-frame semantic interference.

In terms of computational cost, the reorganization is performed once per story and adds an overhead that scales quadratically with the number of frames. 
Let $d$ denote the embedding dimension and $N$ the number of frames; the worst-case cost is $\mathcal{O}(N^2 d)$ up to multiplicative factors that depend on the prompt length (i.e., the number of frame tokens).
This overhead is negligible compared to the diffusion inference cost, and \textit{ReDiStory} requires no backward passes or additional model evaluations.

\begin{figure*}[t] 
    \centering 
    \includegraphics[width=\linewidth]{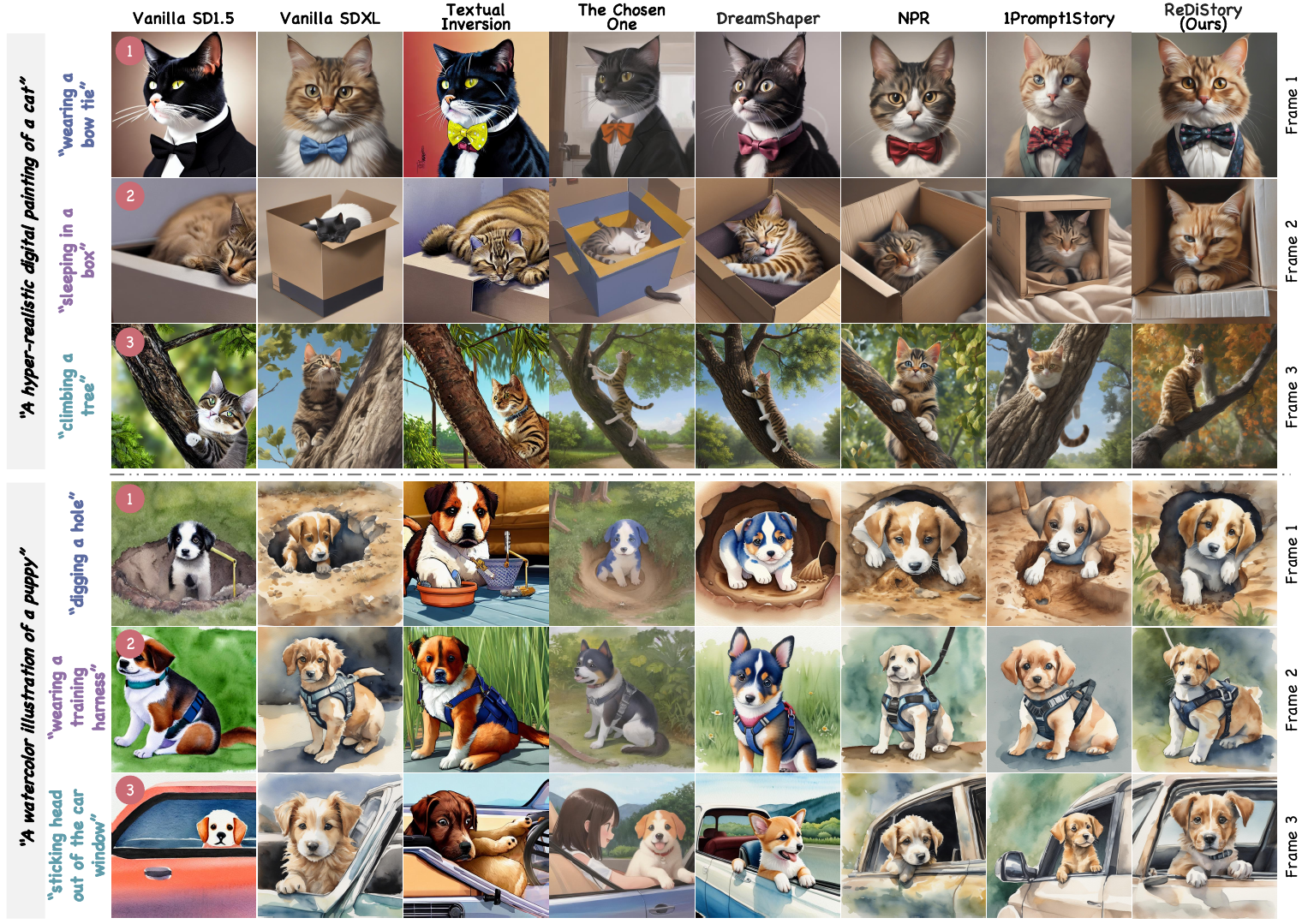} 
    \vspace{-10pt}
    % \caption{Comparison results.} 
    \caption{Qualitative comparison on multi-frame generation. Under identical identity and frame prompts, \textit{ReDiStory} preserves subject-specific appearance as the scene and pose vary.} 
    \label{fig:comparison} 
\end{figure*}

\begin{table*}[t]
\centering
\caption{Quantitative comparison on ConsiStory+. \textbf{Bold} and \underline{underline} denote the best and second best results among \emph{training-free multi-frame} methods, respectively. Vanilla SD1.5 and Vanilla SDXL are shown as model references and excluded from ranking. BLIP-Diffusion and Textual Inversion are based on SD1.5; all remaining methods use SDXL. Although the numerical gains are marginal, ReDiStory consistently improves identity consistency across metrics.}
\resizebox{0.89\linewidth}{!}{
\begin{tabular}{lcccccc}
\toprule
\textbf{Method} & \textbf{CLIP-T$\uparrow$} & \textbf{CLIP-I$\uparrow$} & \textbf{DreamSim$\downarrow$} & \textbf{Steps} & \textbf{Memory (GB)$\downarrow$} & \textbf{Inference Time (s)$\downarrow$} \\
\midrule

\multicolumn{7}{l}{\textit{Vanilla reference models (excluded from ranking)}} \\
\addlinespace[0.2em]
Vanilla SD1.5 & 0.8353 & 0.7474 & 0.5873 & 50 & 4.73  & 2.4657 \\
Vanilla SDXL  & 0.9074 & 0.8165 & 0.5292 & 50 & 16.04 & 13.0890 \\
\addlinespace[0.35em]
\midrule

\multicolumn{7}{l}{\textit{Training-based / reference-guided methods (reported as references)}} \\
\addlinespace[0.2em]
BLIP-Diffusion    & 0.7607 & 0.8863 & 0.2830 & 26 & \textbf{7.75} & \textbf{1.9284} \\
Textual Inversion  & 0.8378 & 0.8229 & 0.4268 & 40 & 32.94 & 282.507 \\
The Chosen One     & 0.7614 & 0.7831 & 0.4929 & 35 & \underline{10.93} & \underline{11.2073} \\
PhotoMaker         & 0.8651 & 0.8465 & 0.3996 & 50 & 23.79 & 18.0259 \\
IP-Adapter         & 0.8458 & \textbf{0.9429} & \textbf{0.1462} & 30 & 19.39 & 13.4594 \\
\addlinespace[0.35em]
\midrule

\multicolumn{7}{l}{\textit{Training-free multi-frame methods}} \\
\addlinespace[0.2em]
ConsiStory                 & 0.8769 & 0.8737 & 0.3188 & 50 & 34.55 & 34.5894 \\
StoryDiffusion             & 0.8877 & 0.8755 & 0.3212 & 50 & 45.61 & 25.6928 \\
Naive Prompt Reweighting (NPR) & 0.8411 & 0.8916 & 0.2548 & 50 & \textbf{16.04} & \textbf{17.2413} \\
1Prompt1Story              & \underline{0.8942} & \underline{0.9117} & \underline{0.1993} & 50 & \underline{18.70} & \underline{23.2088} \\
\textbf{\textit{ReDiStory (Ours)}} & \textbf{0.8966} & \textbf{0.9149} & \textbf{0.1952} & 50 & 18.89 & 23.6413 \\
\bottomrule
\end{tabular}
}
\label{tab:comparison}
\end{table*}

\section{Experiments}
\label{sec:exp}
\vspace{-10pt}
\subsection{Experimental Setup}
\label{sec:setup}

% \noindent\textbf{Dataset.}
\subsubsection{Dataset}
We evaluate \textit{ReDiStory} on the ConsiStory+ benchmark~\cite{liu2025one}, which is a standard testbed for training-free multi-frame storytelling.
The dataset consists of 50 subjects spanning diverse categories, including humans, animals, objects, and fictional characters.
For each subject, 5 frame prompts describe distinct scenes, actions, or viewpoints, resulting in a total of 250 multi-frame test cases.
All methods are evaluated under the same prompt settings to ensure a fair comparison.

\vspace{-10pt}
% \noindent\textbf{Baselines.}
\subsubsection{Baselines}
We compare \textit{DeCorStory} with several representative text-to-image consistency generation methods, including 1Prompt1Story \cite{liu2025one}, BLIP-Diffusion \cite{li2023blipdiffusion}, Textual Inversion (TI) \cite{ruiz2023dreambooth}, IP-Adapter \cite{ye2023ip}, PhotoMaker \cite{li2024photomaker}, The Chosen One \cite{avrahami2024chosen}, ConsiStory \cite{tewel2024training}, and StoryDiffusion \cite{zhou2024storydiffusion}.
All baselines are evaluated using their official configurations or publicly released implementations to ensure fairness.

\vspace{-10pt}
% \noindent\textbf{Implementation details.}
\subsubsection{Implementation details}
All experiments are conducted using SDXL~\cite{podell2023sdxl} implemented with the \texttt{diffusers} library. We use 50 denoising steps and a classifier-free guidance scale of 7.5 for all methods. Unless otherwise specified, we fix the random seed for reproducibility.
% All experiments are performed on a single NVIDIA A100 GPU.
% Additional implementation details and hyperparameters are provided in Appendix~\ref{app:impl}.

% \subsection{Evaluation Metrics}
% \label{sec:metrics}
% \noindent\textbf{Evaluation metrics.}
% We evaluate ReDiStory from two complementary perspectives: prompt fidelity and identity consistency.
% To assess prompt alignment, we compute the average CLIPScore~\cite{hessel2021clipscore} between each generated image and its corresponding frame prompt, which we denote as \textbf{CLIP-T}. A higher CLIP-T score indicates better alignment with the textual description.
% For identity consistency, we measure visual similarity across generated frames using two metrics. Specifically, we adopt \textbf{CLIP-I}, defined as the cosine similarity between CLIP image embeddings~\cite{hessel2021clipscore}, and \textbf{DreamSim}~\cite{fu2023dreamsim}, a perceptual similarity metric shown to closely correlate with human judgments of visual consistency. Following the evaluation protocol of DreamSim, we remove image backgrounds using CarveKit~\cite{selin2023carvekit} and replace them with random noise, ensuring that similarity measurements primarily reflect subject identity rather than background content.

\vspace{-10pt}
% \noindent\textbf{Evaluation metrics.}
\subsubsection{Evaluation metrics}
We evaluate \textit{ReDiStory} from two aspects: prompt fidelity and identity consistency.
Prompt alignment is measured by {CLIP-T}, computed as the average CLIPScore~\cite{hessel2021clipscore} between each generated image and its corresponding frame prompt. Identity consistency is evaluated using {CLIP-I}, defined as the cosine similarity between CLIP image embeddings, and {DreamSim}~\cite{fu2023dreamsim}, a perceptual similarity metric correlated with human judgment.
Following the DreamSim protocol, image backgrounds are removed using CarveKit~\cite{selin2023carvekit} to focus the evaluation on subject identity.

\vspace{-10pt}
\subsection{Comparison}
% \noindent\textbf{Quantitative Results.}
\subsubsection{Quantitative Results}
Table~\ref{tab:comparison} summarizes performance and efficiency. Within training-free methods, \textit{ReDiStory} provides consistent (though numerically modest) improvements in identity consistency over the strongest baseline 1Prompt1Story with higher CLIP-I (0.9149 vs.\ 0.9117) and lower DreamSim (0.1952 vs.\ 0.1993), while maintaining comparable prompt fidelity (CLIP-T). 
% \textit{ReDiStory} also consistently outperforms other training-free baselines, including ConsiStory, StoryDiffusion, and Naive Prompt Reweighting, across all consistency metrics. In addition, \textit{ReDiStory} introduces only marginal computational overhead compared to 1Prompt1Story, with similar memory usage (18.89\,GB vs.\ 18.70\,GB) and inference time (23.64\,s vs.\ 23.21\,s), demonstrating that prompt embedding reorganization remains efficient and practical for inference-time deployment.
This gain comes with a small efficiency trade-off: memory increases from 18.70\,GB to 18.89\,GB and inference time from 23.21\,s to 23.64\,s, indicating that prompt-level reorganization improves consistency with only modest additional cost.

\vspace{-10pt}
% \noindent\textbf{Qualitative Results.}
\subsubsection{Qualitative Results}
Figure~\ref{fig:comparison} shows representative examples. For the same subject, 1Prompt1Story can exhibit gradual identity drift (e.g., subtle changes in appearance details) as frame variations become larger. \textit{ReDiStory} better preserves subject-specific characteristics across frames while still reflecting the frame-level prompts. This suggests that improved consistency does not come at the expense of visual diversity or prompt adherence.

\vspace{-10pt}
\subsection{Ablation Study}
\label{sec:ablation}

Table~\ref{tab:ablation} analyzes the effect of prompt embedding reorganization. Removing reorganization leads to lower identity consistency, indicated by reduced CLIP-I and higher DreamSim. Reorganizing only identity-related embeddings yields an intermediate improvement, while the full \textit{ReDiStory} model achieves the best results across all metrics.
These findings support our hypothesis that decorrelating identity-related and frame-specific semantics is crucial for mitigating inter-frame semantic interference, and that the proposed reorganization improves consistency without degrading prompt fidelity.

% These findings support our hypothesis that mitigating cross-frame coupling at the embedding level is beneficial for identity preservation.

\begin{table}[t]
\centering
\caption{Ablation study on the ConsiStory+ benchmark. Progressively adding embedding reorganization improves identity consistency (higher CLIP-I, lower DreamSim) while keeping prompt fidelity (CLIP-T) stable.}
\vspace{-5pt}
\label{tab:ablation}
\resizebox{\linewidth}{!}{
\begin{tabular}{lccc}
\toprule
\textbf{Method} & \textbf{CLIP-T}$\uparrow$ & \textbf{CLIP-I}$\uparrow$ & \textbf{DreamSim}$\downarrow$ \\
\midrule
w/o reorganization & 0.8921 & 0.9078 & 0.2014 \\
identity-only reorganization & \underline{0.8943} & \underline{0.9112} & \underline{0.1986} \\
\textbf{\textit{ReDiStory} (Full)} & \textbf{0.8966} & \textbf{0.9149} & \textbf{0.1952} \\
\bottomrule
\end{tabular}
}
\end{table}

% \section{Discussion}
% \label{sec:discussion}
% \section{Limitations}
% Despite its effectiveness, \textit{ReDiStory} has several limitations.
% \textbf{(1) Scalability with respect to sequence length.} \textit{ReDiStory} follows the single-prompt formulation, and its performance is therefore bounded by the context length of the text encoder, which limits the number of frames that can be generated in a single pass.
% \textbf{(2) Challenging cases with extreme appearance changes.} Although prompt embedding reorganization improves identity consistency in most scenarios, cases involving severe occlusion or drastic viewpoint changes may still pose challenges, as identity cues become less reliable under such conditions.

% \noindent\textbf{Computational Cost.}
% ReDiStory introduces negligible computational overhead compared to standard inference pipelines, as it does not modify the diffusion model, introduce additional optimization objectives, or require extra forward or backward passes. The primary additional cost lies in prompt construction and input preprocessing, which is insignificant relative to the overall generation time. As a result, ReDiStory maintains the efficiency and practicality of training-free inference-time methods.

\vspace{-10pt}
\section{Conclusion}
\label{sec:conclusion}

We presented \textit{ReDiStory}, a training-free framework for multi-frame visual story generation that improves subject consistency via inference-time prompt embedding reorganization. By separating identity-related and frame-specific embeddings and suppressing shared directions across frames, ReDiStory mitigates cross-frame semantic interference without modifying the diffusion backbone. Under identical settings, this prompt-level intervention consistently improves identity preservation with minimal computational overhead. While the gains are modest, the results indicate that reducing cross-frame coupling at the embedding level is an effective and lightweight approach, suggesting future work on strengthening the effect for longer sequences and more challenging appearance variations.

\clearpage
\bibliographystyle{IEEEbib}
\bibliography{main2}

\clearpage
\appendix
\onecolumn

\setcounter{page}{1}
% \maketitlesupplementary
\setcounter{section}{0}
\setcounter{figure}{0}
\setcounter{table}{0}

\renewcommand{\thesection}{A.\arabic{section}}
\renewcommand{\thefigure}{A.\arabic{figure}}
\renewcommand{\thetable}{A.\arabic{table}}
% \renewcommand{\thelisting}{A.\arabic{listing}}

% \section{Additional Experimental Details}
% \label{app:impl}

% \subsection{Implementation Details}
% We report additional implementation details, including random seeds, hardware configuration, library versions, and hyperparameters.

% \subsection{Additional Quantitative Results}
% This section provides per-category results on the ConsiStory+ benchmark.

% \subsection{Additional Qualitative Results}
% We include additional qualitative examples and failure cases.

\clearpage
\section{Appendix}

\begin{figure}[h]
    \centering
    \includegraphics[width=1\linewidth]{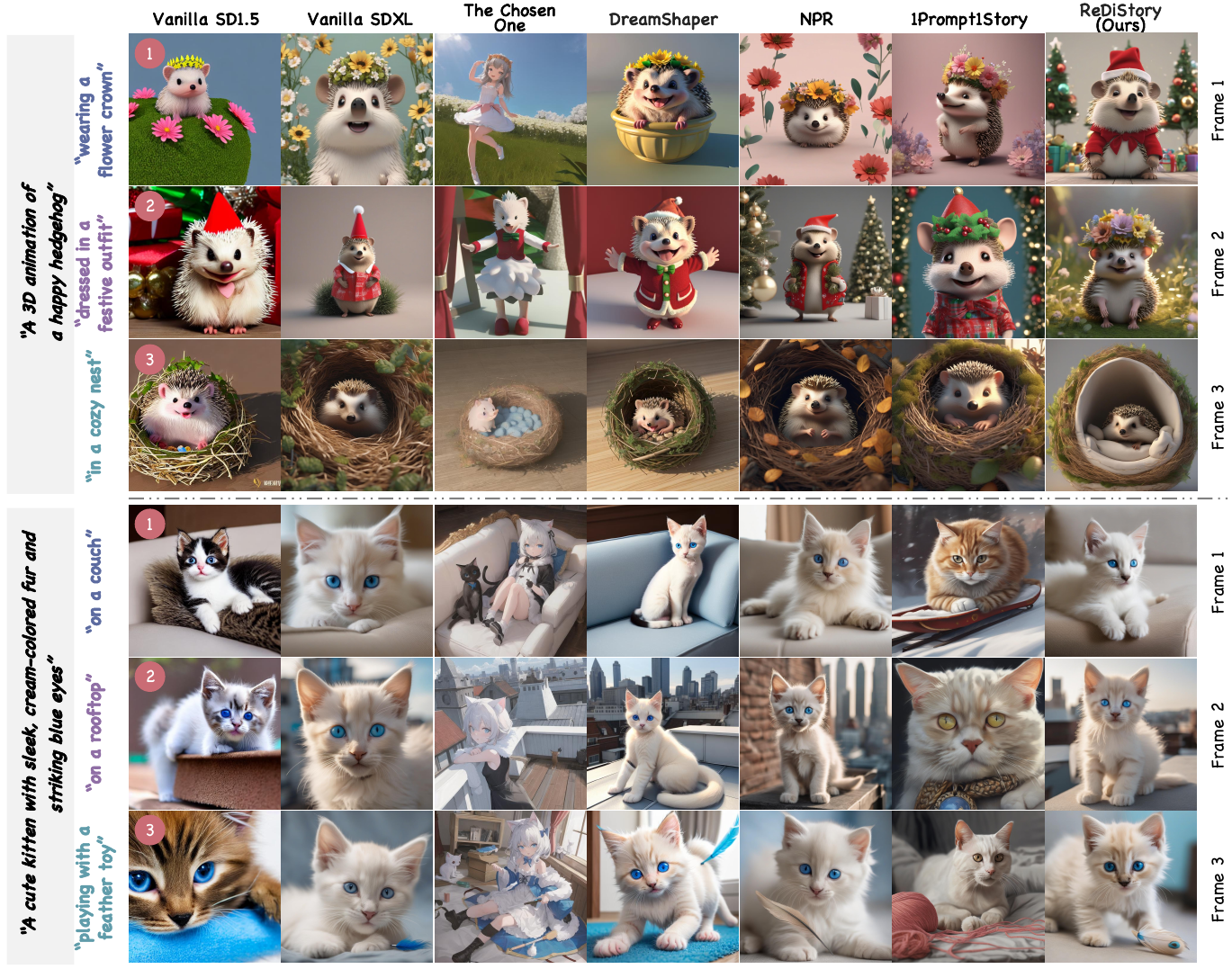}
    \caption{Additional comparison results.}
    \label{fig:comparison_appendix}
\end{figure}

\end{document}